# An Improved Approach for Word Ambiguity Removal

**Priti Saktel**  *saktel.priti10@rediffmail.com*
*Department of Computer Science and Engineering*
*G. H. Raisoni College of Engineering*
*Nagpur, MS, 440009, INDIA*

**Urmila Shrawankar**  *urmila@ieee.org*
*Department of Computer Science and Engineering*
*G. H. Raisoni College of Engineering*
*Nagpur, MS, 440016, INDIA*

**Abstract**

Word ambiguity removal is a task of removing ambiguity from a word, i.e. correct sense of word is identified from ambiguous sentences. This paper describes a model that uses Part of Speech tagger and three categories for word sense disambiguation (WSD). Human Computer Interaction is very needful to improve interactions between users and computers. For this, the Supervised and Unsupervised methods are combined. The WSD algorithm is used to find the efficient and accurate sense of a word based on domain information. The accuracy of this work is evaluated with the aim of finding best suitable domain of word.

## 1. INTRODUCTION

Sometimes people are facing problems in understanding correct meaning of the sentence. Since, sentence comprised of ambiguous words. In such case, correct meaning is taken by the context of the sentence. Usually, it is found in English language. In other words, we can say that context uniquely identifies meaning of the sentence. Based on this interpretation the ambiguity of word, known as lexical ambiguity is disambiguated; which is called as a process of WSD. Manual method of meaning extraction uses approach of searching words correct meaning in typical or online dictionaries which had several drawbacks.

To resolve an ambiguity in a sentence, natural language processing provides word sense disambiguation which governs a sentence in which the sense of a word or meaning is used, when the word has multiple meanings (polysemy). WSD is a process which identifies the correct sense of a word with the help of surrounding words in a sentence. The correct sense of a word is obtained from the context of the sentence. a different meaning of the single word is associated in each sentence based on the context, the remaining sentence gives us. Thus, if the word imagination appears near the word play, we can say that it is related to free_time and not related to a sport which is known as local context. Computers that read words, one at a time must use word sense disambiguation process for finding the correct meaning (sense) of a word. A disambiguation process requires a dictionary in which senses are to be specified and disambiguated. For identifying the correct sense of the word the 'WordNet' domain is used. A domain consists of different syntactic categories of synsets. It groups senses of the same word into uniform clusters, with the effect of reducing word polysemy in WordNet. WordNet domain provides semantic domain as a natural way to establish semantic relations among word senses. This functionality is used in creation of MySQL database. The system for disambiguation of ambiguity in a sentence aims to identify domain of intended sense of word. Basically, input provided to the system is a sentence with ambiguous words and the output is identified as domain of word.

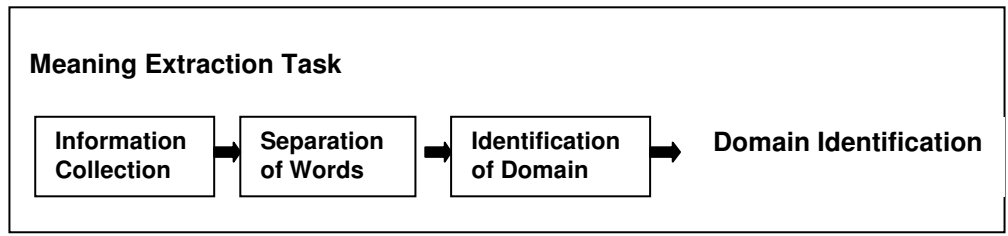

**FIGURE 1** Context based meaning extraction task

## 2. LITERATURE SURVEY

For Word sense disambiguation, the first attempt effectively used by Michael E. Lesk was based on the Dictionary approach [1]. The problem with this algorithm is that, it defines context in a more complex way which is overcome by Simplified Lesk algorithm [2]. It can be effectively used with the WordNet lexical database. Such an attempt is made at Indian Institute of Technology, Bombay [3] and the results are promising. Navigili [4] had found that the right sense for a given word amounts to identifying the most "important" node among the set of graph nodes representing its senses. Ling Che Yangsen and Zhang [5] described a general framework for domain adaptation which contained instance pruning and weighting and the training instance augmentation. Agirre [6] described a thorough overview of the current WSD techniques and performance of systems on data sets, as well as a brief history of the field and some truly insightful discussions on potential developments. In [7] we find the most general and well-known attempt to utilize information in machine-readable dictionaries for WSD, that of Lesk , which computes a degree of overlap—that is, number of shared words--in definition texts of words that appear in a ten-word window of context.

## 3. SYSTEM MODEL

The system model has five stages:
**POS Tagger**
An English sentence with ambiguous words is given as an input to the project. From the sentence, content words are extracted and tagged by POS tagger [6, 7, 8] [22].
**Distribute Domain**
Then domains are distributed to Content words from the WordNet Domains which maintains domain distribution table [3, 4, 5] [22].
**Pick the Target Word**
The target word is selected by comparing WordNet, available domain and the domain of target word is displayed.
**Identification of Domain**
The accurate domain of the target word is identified by supervised and unsupervised training [1] [2] [22].
**Obtain Sense of Word**
The sense of target word belonging to the domain is obtained which is added to the domain distribution table i.e. the table is updated using supervised and unsupervised training [4].

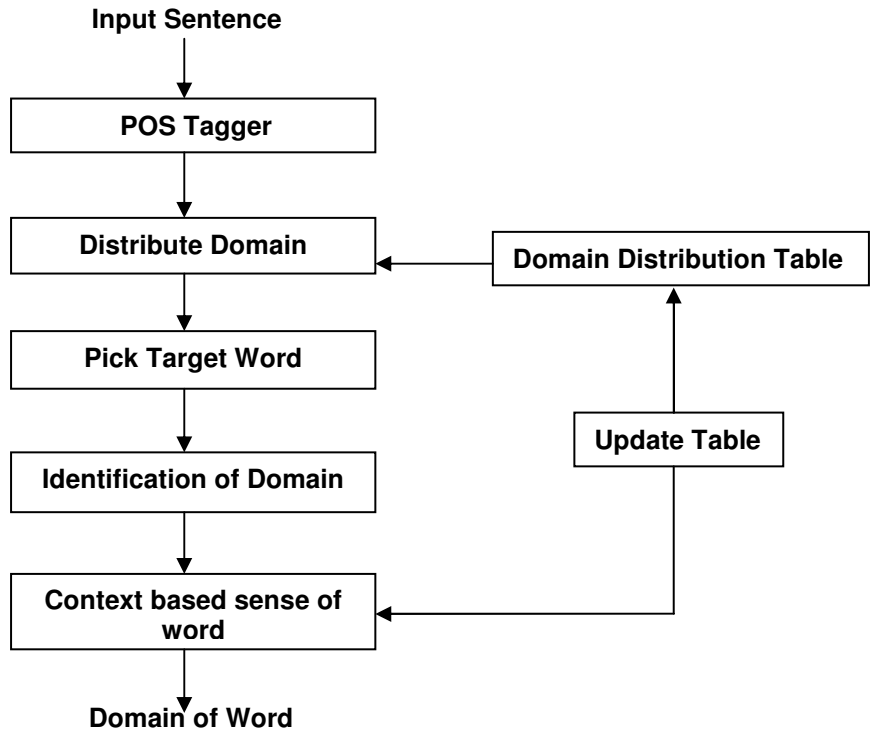

**FIGURE 2** System Model

## 4. WSD ALGORITHM

This algorithm is used in supervised and unsupervised training method and gives better performance than graph based algorithm. [13].It has following steps:
**Step1:** Create a database which can store the words and their meanings.

| ID | field | ID | Word | ID | Word | FieldID |
|---|---|---|---|---|---|---|
| 1 | Computer | 70 | is | 441 | diving | 2 |
| 2 | Sports | 71 | the | 442 | racing | 2 |
| 3 | Medical | 72 | was | 443 | athletics | 2 |
| 4 | Engineering | 73 | that | 444 | wrestling | 2 |
| 5 | Factotum | 74 | on | 445 | boxing | 2 |
| 6 | History | 75 | of | 446 | fencing | 2 |
| 7 | Geography | 76 | for | 447 | archery | 2 |
| 8 | Games | 77 | where | 448 | fishing | 2 |
| 9 | Law | 78 | how | 449 | hunting | 2 |
| 10 | Biomedical | 79 | when | 450 | bowling | 2 |

**TABLE 1:** Fields Table **TABLE 2:** General Words Table **TABLE 3:** Meanings Table

The three tables are created as fields, general words and meanings. TABLE 1 shows fields table in which ID and Domain name is stored. An ID is assigned to respective domain name. TABLE 2 shows General words table in which ID and general words are stored after separation of words. TABLE 3 shows Meanings table in which ID, words and respective domain ID assigned to words are stored. A unique ID and FieldID are assigned to the word which belonging to correct domain name.

**Step 2:** Separate the content words from the sentence using Part -of- Speech tagging (POS) process. This process is used for identification of words as nouns, verbs, adjectives, adverbs, etc, since it is used to tag or mark the text [11]. FIGURE 3 shows tags which are used to mark the content words and their separation. The separation is done with the help of Penn Treebank Tagset of Part of Speech tagging process which is shown in FIGURE 3 and FIGURE 4.

Example:

**Play the stock market.**

| Tag | Description(Penn Treebank Tagset) |
|---|---|
| DTR | Determiner |
| NN | Noun,Singular or mass |
| VBD | Verb,Past tense |
| VBG | Verb present participle |
| NNS | Noun,plural |

**FIGURE 3** Penn Treebank Tagset

The |DTR fisherman |VBD went |VBG to |the |NN bank |NN.|.

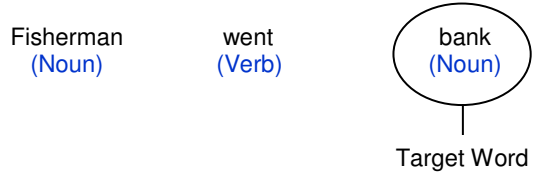

Fisherman (Noun)    went (Verb)    bank (Noun)

Target Word

**FIGURE 4** Example of POS Tagger Process

**Step 3:** Decompose the separation of sentence into three categories as C1, C2 and C3 for finding results i.e. displaying correct domain of word. In step 3, various comparisons are performed to find correct domain of words. It is required to detect correct sense of word with the help of most suitable domain for a word using various algorithms and finally the meaning of a sentence.

| Fisherman (noun sense) Went (verb sense) Bank (noun sense) | Fisherman (noun sense) — Profession | Went (verb sense) — Factotum | Bank (noun sense) — Factotum Economy Nature | Factotum Economy Nature |
|---|---|---|---|---|
| C1 | | C2 | | C3 |

**FIGURE 5** Contents of Category C1, C2 and C3

**Step 4:** Supervised training module to check if the given category of words are properly processed or not. In step 2, if the inputted sentence domain displayed by the system is free_time. But this may be a wrong domain if the context based meaning is considered. According to the context, domain of play is Commerce. Since stock market is whose work is related to Commerce.

Priti Saktel & Urmila Shrawankar

In this case, supervised training is required to train the system to pick the correct domain as Commerce. Let us assume that the sentence is
The play of the imagination.
The correct domain for the word play is Free_time. Since maximum count of comparison is 2 for domain free_time (ID 4).Suppose the next sentence is entered by user is
Play the drama.
Here, the domain of the word play and drama is Entertainment. Previously, the same word has domain related to free_time [10, 12]. It is shown in TABLE 4 below.

| FieldID | Word | Domain |
|---|---|---|
| 4 | Play | Free_time |
| 5 | play | commerce |
| 4 | imagination | Free_time |

**TABLE 4** Domain Comparisons

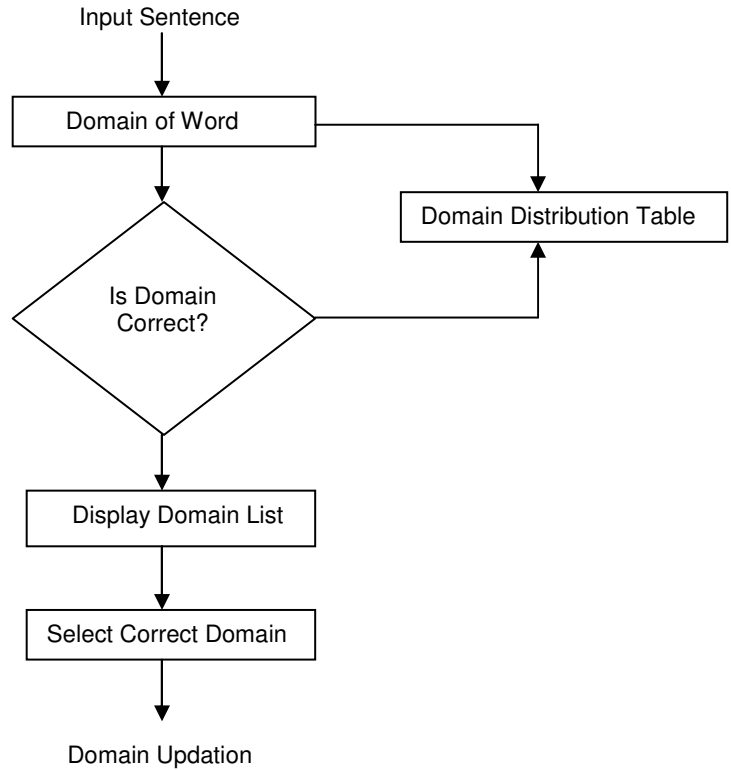

**FIGURE 6** Supervised Training Flow

**Step 5:** Unsupervised learning module to auto update the database with the selected sentences and word-meaning pairs. The flow is shown in fig. 7. If it is correct that is considered as correct domain of word (disambiguation) and this entry is updated in the database. Else, user has given the chance to input the sentence again. This flow is shown in Fig. 7. The knowledge acquisition bottleneck problem is overcome by unsupervised learning, since it is independent of manual work.

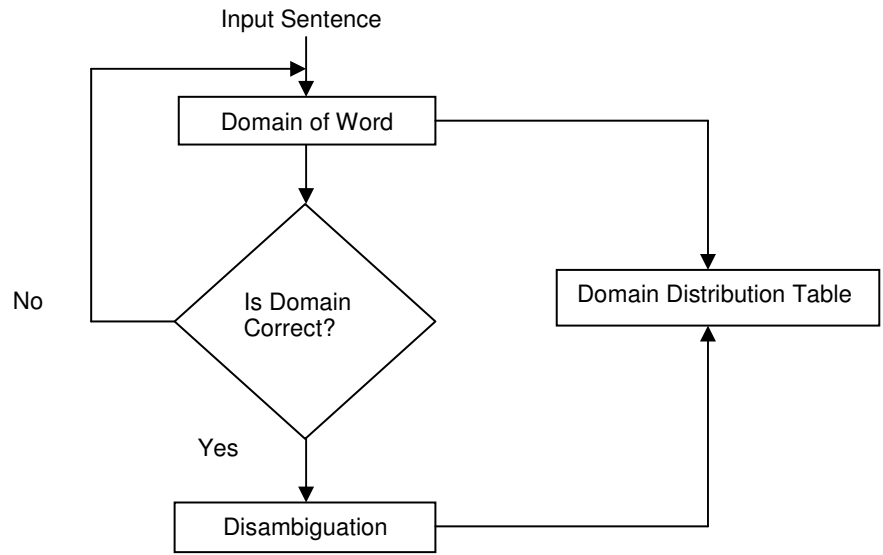

**FIGURE 7** Unsupervised Learning Flow

The experimental setup is done by following steps and accuracy of Unsupervised, Supervised and Hybrid method is evaluated using mathematical formula as

$$\frac{\sum_t \text{Number of Correct terms}}{\sum_i \text{Number of Input}}$$

Where,  t = correct terms (Correctly disambiguated)
&  i = input (Number of sentences)

Repeat the below steps for:
   i=1…number of sentences (n), n=1….15
   Where,  i indicate sentence and n indicates number of sentences.

**Step 6:** Finally, display the **correct** domain of the word. The correct domain of the word for given example is Commerce.

## 5. RESULTS

**Stage 1: Part of Speech Tagger**
The first stage "POS Tagger" of the system model is implemented. FIGURE 9 shows the snapshot of POS Tagger process. This stage is used to separate the content words and general words like noun, verb, adjective etc. from the sentence in step1 and Classification of separated words in three categories c1, c2, and c3.

Priti Saktel & Urmila Shrawankar

> **Sentence-** Play the stock market.
> **Separation-**
> Play
> The
> Stock
> Market
> Match: play: play clustered under – Commerce
> Match: play: play clustered under – Free_time
> Match: play: play clustered under – Entertainment
> Match: stock: stock clustered under – Commerce
> Match: market: market clustered under - Commerce

**FIGURE 9** Result of POS Tagger Process

**Stage 2: Unsupervised Learning**
There are five steps to process the system. When the domain of word is identified; it is checked by the system for correctness if the identified domain is correct then out of five steps only four steps are processed to get the output. This is shown in FIGURE 10 with example.

> **Sentence: The play of the imagination.**
>
> **Step 1: Separating All Words**
> Word: The
> Word: play
> Word: of
> Word: the
> Word: imagination.
>
> **Step 2: Finding Matching Domain**
> Match – play: play
> Match – play: play
> Match – play: play
> Match – play: play
> Match – play: play
> Match – imagination: imagination
> Match – imagination: imagination
>
> **Step 3: Checking for Best Probable Field**
> Field 11 found 2 times
> Field 2 found 2 times
> Max Value: 9 For field ID: 69
> The Domain is Free_time
>
> **Step 4: Checking for Correctness**
> Is this the type of the sentence at input? Y/N
> The new elements with selected domains have been updated…

**FIGURE 10** Result of Unsupervised Learning of Implemented System

Priti Saktel & Urmila Shrawankar

---

**Sentence: Play the stock market.**
**Step 1: Separating All Words**
Word: Play
Word: the
Word: stock
Word: market.
**Step 2: Finding Matching Domain**
Match – play: Play
Match – play: Play
Match – play: Play
Match – play: Play
Match – play: Play
Match – stock: stock
Match – stock: stock
Match – market: market
Match – market: market
**Step 3: Checking for Best Probable Field**
Field 11 found 1 times
Field 17 found 3 times
Field 2 found 2 times
Field 69 found 9 times
Max value: 9 for field ID: 69
The Domain is Free_time
**Step 4: Checking for Correctness**
Is this the type of the sentence at input? Y/N
**Step 5: Supervised Learning**
Your choice is: **Commerce**
So the new field of this sentence is set to: Commerce
Words to be updated:

**FIGURE 11** Result of Supervised Learning of Implemented System

**Stage 3: Supervised Learning**
In stage 3, when the domain of word is identified; it is checked by the system for correctness if the identified domain is incorrect then all five steps of system are processed to get the output. This is shown in FIGURE 11 with example.

**Stage 4: Spell Checker Utility**
Sometimes, the sentence entered by the user will be incorrect or correct. So, here apart from above results one additional step as spell checker utility is implemented. In stage 4, the corrections in spellings of the entered sentence are corrected using online spell checker concept which requires internet connection before executing the system. The result of this utility is shown below in FIGURE 12.

---

Pla the stk makt.

Probable Spelling Matches found...

Play    Ply    Plum

stuck    stock    stick    wore    worm

marketing    market    making

Do you wish to change the input(y/n): y

---

**FIGURE 12** Result of Spell Checker utility

International Journal of Human Computer Interaction (IJHCI), Volume (3) : Issue (3) : 2012          78

Priti Saktel & Urmila Shrawankar

**Stage 5: Final Result of the System**
The system is used for determining correct domain of word. First part of this system is sentence collection. It is required by the user to enter the sentence after that sentence is separated by POS tagger. Once the sentence is separated out, it will be processed through various steps like domain distribution, supervised learning, unsupervised learning, WSD algorithm. The final result for the implemented system is shown below in TABLE 5:

| Sentence | Separation of Words | Target Word | Domain Identification | Comparison | Final Domain |
|---|---|---|---|---|---|
| Play the stock market | Play the stock market | Match – play: play  Clustered under Match –stock: stock  Clustered under Match – market: market  Clustered under Match –play: play | Entertainment  Commerce  Commerce  Commerce | Max Value :03 For field ID: 05 | Commerce |

**TABLE 5** Final Result of Implemented System

**Stage 6: Results of Accuracy of the System**
Firstly, the unsupervised learning, supervised learning and hybrid training accuracy is evaluated shown in TABLE 6 and FIGURE 13.Then comparison of all learning approaches are done and observed that these approaches gives 63%,76% and 80% of accuracy respectively. Hence, the accuracy is improved using Hybrid training method shown in TABLE 7 and FIGURE 14.

| Sentence | Target word | Disambiguated | Correctly Disambiguated | Accuracy (%) |
|---|---|---|---|---|
| 1 | 2 | 2 | 2 | 100 |
| 2 | 3 | 3 | 2 | 66.67 |
| 3 | 1 | 1 | 1 | 100 |
| 4 | 1 | 1 | 1 | 100 |
| 5 | 2 | 2 | 1 | 50 |
| 6 | 2 | 2 | 2 | 100 |
| 7 | 2 | 2 | 1 | 50 |
| 8 | 1 | 1 | 1 | 100 |
| 9 | 1 | 1 | 1 | 100 |
| 10 | 2 | 1 | 1 | 100 |
| 11 | 3 | 3 | 2 | 66.67 |
| 12 | 1 | 1 | 1 | 100 |
| 13 | 3 | 3 | 2 | 66.67 |
| 14 | 1 | 1 | 1 | 100 |
| 15 | 2 | 1 | 1 | 50 |
| Total | 27 | 25 | 20 | 80.00 |

**TABLE 6** Results of Hybrid Learning Method Accuracy of 15 Sentences

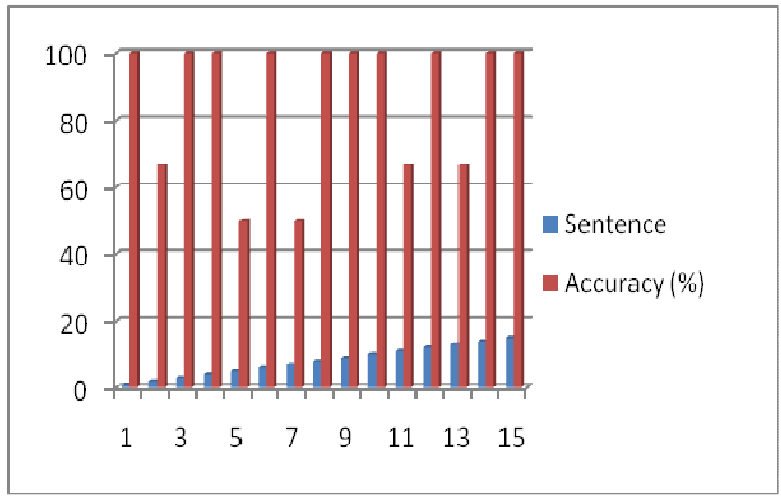

**FIGURE13** Hybrid Learning Method Accuracy

| Sentence | Target word | Disambi-guated | Correctly disambiguated | Supervised Accuracy (%) | Unsupervised Accuracy (%) | Hybrid Accuracy (%) |
|---|---|---|---|---|---|---|
| 1 | 2 | 2 | 1 | 50 | 100 | 100 |
| 2 | 3 | 3 | 2 | 67 | 67 | 67 |
| 3 | 1 | 1 | 1 | 100 | 100 | 100 |
| 4 | 1 | 1 | 1 | 100 | 100 | 100 |
| 5 | 2 | 2 | 1 | 50 | 50 | 50 |
| 6 | 2 | 2 | 2 | 100 | 100 | 100 |
| 7 | 2 | 2 | 1 | 50 | 50 | 50 |
| 8 | 1 | 1 | 1 | 100 | 100 | 100 |
| 9 | 1 | 1 | 1 | 100 | 100 | 100 |
| 10 | 3 | 3 | 1 | 33 | 50 | 50 |
| 11 | 3 | 3 | 2 | 67 | 67 | 67 |
| 12 | 2 | 2 | 1 | 50 | 100 | 100 |
| 13 | 3 | 3 | 2 | 67 | 67 | 67 |
| 14 | 3 | 2 | 1 | 50 | 100 | 100 |
| 15 | 2 | 2 | 1 | 50 | 50 | 50 |
| Total | 31 | 30 | 19 | 63 | 76 | 80 |

**TABLE 7** Results of Comparison of Unsupervised, Supervised and Hybrid Learning

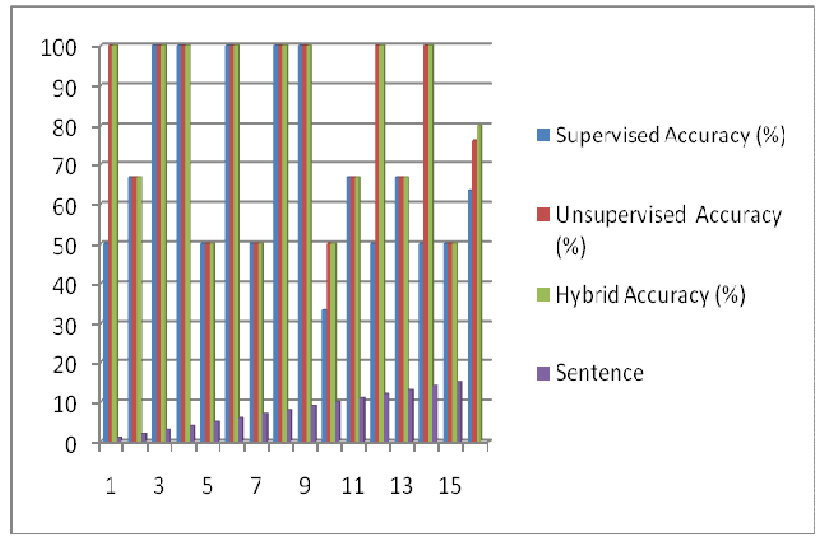

FIGURE 14 Comparison of Unsupervised, Supervised, Hybrid Accuracy

## CONCLUSIONS

The system improves the self-learning process by obtaining correct sense of a sentence by resolving ambiguity from a word with full automation. The system requires correct domain of word identification from the sentence. Hence, sentence comprised of various content words like nouns, verbs, adjectives, adverb etc. Firstly, it is required to separate out content words from a sentence. By applying POS tagger process and WSD algorithm, domain is allotted to each word and each domain of word is compared to get correct domain of word. A count of comparisons is calculated, the domain which has the maximum count is assumed as correct domain. Also, this system improves the accuracy of identifying the correct domain of word. As per the Table 8 it shows that self learning language is improved by obtaining correct sense of a word by removing ambiguity from a sentence with full automation. Also, improves disambiguation process by obtaining appropriate sense of a word. The synonym relationship approach is used to identify intended domain of word. The system is trained using supervised training to check correctness of domain which gives 76% of accuracy; an unsupervised learning is used to update the database with the selected sentences and word-meaning pairs automatically. It gives 63% of accuracy. The hybrid method improves this accuracy up to 80% from Table 7. In this system, when the number of target word is correctly disambiguated system gives 100% accuracy. Else, the accuracy may be 66% or 50%. Hence, the overall 80% accuracy is evaluated. These results generated by the system are beneficial for Human Computer Interaction as it is motivating people to learn the language by themselves using computer in the absence of teacher. Additionally, the spell checker utility is implemented to avoid mistakes in words.

## REFERENCES

[1] Roberto Navigli, Mirella Lapata (2010), "An Experimental Study of graph connectivity for unsupervised word sense disambiguation", IEEE Transactions on Pattern Analysis and Machine Intelligence, Vol. 32, No. 4.
[2] Myunggwon Hwang, Chang Choi (2011)," Automatic Enrichment of Semantic Relation Network and Its application to Word sense Disambiguation",IEEE transaction on knowledge and data engineering, vol. 23, no. 6.

[3] Francisco Tacoa, Hiroshi Uchida (2010),"A Word Sense Disambiguation approach for converting Natural Language Text into a Common Semantic Description", Fourth International Conference on Semantic Computing, IEEE.
[4] Ping Chen, Wei Ding (2010), "Word Sense Disambiguation with Automatically Acquired Knowledge", IEEE Intelligent Systems.
[5] Ling Che Yangsen Zhang (2011),"Study on Word Sense Disambiguation Knowledge base based on Multi-Sources", IEEE.
[6] Alex Roney Mathew; Al Hajj (2011), "Human-Computer Interaction (HCI): An overview", IEEE International Conference on Computer Science and Automation Engineering (CSAE).
[7] E. Agirre and P. Edmonds (2006)," Word sense Disambiguation: Algorithms and Applications", Springer.
[8] Johan Bos and Malvina Nissim (2009)," From shallow to deep Natural language processing: A hands-on tutorial", Springer.
[9] Leung (2006),"Learners as users, and users as learners", 7$^{th}$ International Conference on Information Technology Based Higher Education and Training, ITHET '06.
[10] Yousif, J.H. Sembok, T.(2008),"Arabic part-of-speech tagger based Support Vectors Machines", Information Technology,2008. ITSim 2008. International Symposium.
[11] Diana McCarthy, Rob Koaling (2007),"Unsupervised Acquisition of Predominant Word Senses," Computational Linguistics, Vol.33, No.4.
[12] Andrei Mincă, Stefan Diaconescu (2011),"An Approach to Knowledge-Based Word Sense Disambiguation Using Semantic Trees Built on a WordNet Lexicon Network",IEEE.
[13] Priti Saktel,Urmila Shrawankar(2012),"Context based Meaning Extraction for HCI Using WSD Algorithm:A Review", IEEE-International Conference on Advances in Engineering, Science and Management,pp. 208-212.
[14] Jerome R. Bellegarda, Fellow (2010)," Part-of-Speech Tagging by Latent Analogy", IEEE Journal of Selected Topics In Signal Processing, Vol. 4, No. 6, Dec. 2010.